\documentclass[graybox]{svmult}

\usepackage{type1cm}        
\usepackage{hyperref} 
\usepackage{makeidx}         
\usepackage{graphicx}        
\usepackage{multicol}        
\usepackage[bottom]{footmisc}

\usepackage{newtxtext}       %
\usepackage[varvw]{newtxmath}       

\usepackage{threeparttable}  
\usepackage{booktabs}
\usepackage{multirow}
\usepackage{bm}
\usepackage{pifont}
\newcommand{\cmark}{\ding{51}}%
\newcommand{\xmark}{\ding{55}}%

\makeindex             

\begin{document}

\title*{Can GPT Improve the State of Prior Authorization via Guideline Based Automated Question Answering?}

\author{Shubham Vatsal, Ayush Singh and Shabnam Tafreshi}


\institute{Shubham Vatsal, Ayush Singh and Shabnam Tafreshi \at inQbator AI at eviCore Healthcare, Evernorth Health Services\\ \email{firstname.lastname@evicore.com}}
%
%
\maketitle

\abstract*{Health insurance companies have a defined process called prior authorization (PA) which is a health plan cost-control process that requires doctors and other healthcare professionals to get clearance in advance from a health plan before performing a particular procedure on a patient in order to be eligible for payment coverage. 
For health insurance companies, approving PA requests for patients in the medical domain is a time-consuming and challenging task. One of those key challenges is validating if a request matches up to certain criteria such as age, gender, etc. In this work, we evaluate whether GPT can validate numerous key factors, in turn helping health plans reach a decision drastically faster. We frame it as a question answering task, prompting GPT to answer a question from patient electronic health record. We experiment with different conventional prompting techniques as well as introduce our own novel prompting technique. Moreover, we report qualitative assessment by humans on the natural language generation outputs from our approach. 
Results show that our method achieves superior performance with the mean weighted F1 score of 0.61 as compared to its standard counterparts.\keywords{large language model, GPT, question answering, prior authorization, healthcare}}

\abstract{Health insurance companies have a defined process called prior authorization (PA) which is a health plan cost-control process that requires doctors and other healthcare professionals to get clearance in advance from a health plan before performing a particular procedure on a patient in order to be eligible for payment coverage. 
For health insurance companies, approving PA requests for patients in the medical domain is a time-consuming and challenging task. One of those key challenges is validating if a request matches up to certain criteria such as age, gender, etc. In this work, we evaluate whether GPT can validate numerous key factors, in turn helping health plans reach a decision drastically faster. We frame it as a question answering task, prompting GPT to answer a question from patient electronic health record. We experiment with different conventional prompting techniques as well as introduce our own novel prompting technique. Moreover, we report qualitative assessment by humans on the natural language generation outputs from our approach. 
Results show that our method achieves superior performance with the mean weighted F1 score of 0.61 as compared to its standard counterparts.\keywords{large language model, GPT, question answering, prior authorization, healthcare}}

\section{Introduction}
\label{intro}
Clinical guidelines are an important tool in the PA workflow in helping improve patient outcomes
\footnote{https://www.evicore.com/insights/can-ready-access-to-evidence-based-guidelines-improve-patient-outcomes}.
The guidelines serve as a reference for healthcare professionals to make decisions about the most appropriate path of care for a patient. The guidelines cover a vast array of patient condition scenarios, such as headache, lower back pain, imaging, etc. Furthermore, they are synthesized by subject matter experts based on latest medical literature, assessing the advantages and disadvantages of diagnostic techniques and therapies. These guidelines
\footnote{https://www.nccih.nih.gov/health/providers/clinicalpractice}
provide support to help providers continue to practice evidence-based medicine in an ever-faster innovation environment. The initial development of these guidelines typically involves incremental steps
\footnote{https://www.evicore.com/insights/provider-clinical-guidelines}.
These guidelines do not have the same legal weight as directives. In other words, if a doctor doesn't think a recommendation is appropriate for a particular patient, they are not obligated to follow it. However, exceptions to rules need to be explained. 

Even though guidelines list the key criteria useful for decision making, practitioners nevertheless have to peruse the lengthy patient records to assess which criteria were met or unmet. Moreover, a patient record can have anywhere from 20 - 400 pages of information \cite{apathy2023notebloat}. Reasoning over numerous criteria on a lengthy document is all the more burdensome. Therefore, any effort to alleviate the need to seek the guideline criteria in lengthy records improves clinician's workflow and as a result the state of healthcare overall.

Large language models (LLMs) have gained popularity in the recent past for their effectiveness in reasoning where traditional computational methods failed \cite{thirunavukarasu2023large,vatsal2024survey}. This opens doors to being able to do complex computational reasoning to arrive at answers, which is indispensable in case of guideline-based question answering. Although LLMs have been shown to perform well on well-curated biomedical datasets, their effectiveness on noisy real-world data sets has been sparsely studied. In this work, we evaluate GPT's \cite{openai2023gpt4} capabilities in answering some of the criteria commonly found in clinician's decision-making workflows.

We worked with clinicians in identifying some of the most frequently used guidelines in their workflows. Based on their feedback, requesting spine imaging is a frequent practice in the health insurance domain, where numerous criteria need to be evaluated before a PA request is approved. Therefore, we use the criteria listed in \textit{Spine} guidelines of a major PA provider.
We convert a given clinical guideline criterion to a query known widely as a \textit{prompt}, followed by asking GPT to answer the query from a patient's health records. We identify three major advantages of having GPT seek and validate guidelines as follows:

\begin{itemize}
\item{Increased transparency with an automatically compiled trail of evidence for a decision}
\item{Shorter PA response times due to lesser manual labor by clinicians}
\item{Reduction in inclusion of personal bias of clinicians due to structured question-answering}
\end{itemize}

We begin our evaluation with state of the art prompting approaches of GPT, however, further analysis reveal few drawbacks of existing approaches discussed in section \ref{RW}. To that end, we introduce a novel prompting method named \textit{\textit{Implicit Retrieval Augmented Generation (RAG)}} that first fetches segments of document that are relevant to the query before reasoning over them. Moreover, our approach also provides reasoning behind it's answers. Experiments indicate our approach outperforms other methods by a moderate margin. 
Although machine evaluation is a good measure of performance, it falls short when evaluating artificially generated text \cite{schluter-2017-limits}, where actual human preferences are significantly superior. Therefore, we report qualitative preference metrics by human experts on the output of our proposed approach. We find that humans agree with the generated outputs most of the time.

\section{Problem Statement}
\label{sec:2}
Our overall objective is to mimic the behavior of clinicians who carefully review the underlying patient case for PA. We task GPT with extracting key information from the unstructured text and answering a set of guideline questions. Thus, we frame our problem to one of evaluating GPT's performance as a QA natural language processing system on patient health records. Additionally, clinicians often seek specific parts of the document to answer a question. Along a similar vein, we leverage the generation capabilities of GPT to retrieve segments of documents that it found useful in arriving at an answer. 

\section{Related Work}
\label{RW}
Even though there has been research done around the use of natural language processing or machine learning in the domain of healthcare, albeit seldom in context of PA. \cite{junior2019study} studies the influence of textual features in healthcare PA process. 
\cite{prasanna2021sentiment} talks about sentiment analysis of tweets related to PA.
\cite{choudhury2021using} used machine learning for early prediction of the post-acute care discharge to minimize delays caused by PA. \cite{sk2023taxonomy} created a taxonomy and used it in downstream tasks. 
\cite{de2023determining} works on determining PA approval for Lumbar Stenosis surgery with machine learning. \cite{farias2019using} uses a machine learning approach which incorporates historical patient data to improve PA learning process. 
\cite{ginjupalli2022digitization} aims to digitize the PA process by implementing classification algorithms.

Recently, there has been a surge in research in using generative models like GPT for question-answering in the medical field \cite{thirunavukarasu2023large}.
\cite{kasai2023evaluating} evaluates GPT and ChatGPT on Japanese medical licensing examinations.
\cite{nori2023capabilities, kasai2023evaluating} presents a comprehensive evaluation of GPT-4 on medical competency examinations and benchmark datasets. Chain-of-Thought
(CoT) prompting \cite{wei2022chain} gained popularity because of its ability to extract reasoning along with answer. More recently, \cite{yasunaga2023large} introduced prompting via analogical reasoning which requests the model to do global knowledge recall of labeled exemplars before arriving at an answer. 

While the above mentioned prompting experiments improves performance, it still is limited to only operating from model's internal knowledge. This is where retrieval augmented generation (RAG) helps by providing an external memory store to the model, thereby supplementing it's internal representations \cite{lewis2020retrieval}. \cite{lievin2022can} investigates diverse prompt variations (zero-shot, few-shot, domain-specific CoT cues and retrieval augmentation) for GPT-3.5 on multiple-choice medical board exam question datasets, albeit on real-world healthcare datasets. Our work builds upon the novelty of RAG in a QA setup with the aim of speeding-up and increasing accuracy of PA processes. 

\section{Dataset}
Our dataset comprises of patient case reports or health records that contain unstructured text. There are tens of thousands of clinical guidelines and answering questions corresponding to all the guidelines, however, running them via GPT would be cost-prohibitive. Therefore, we worked with clinicians in identifying some of the most frequently used guidelines in their workflows.
Based on their feedback, requesting spine imaging is a frequent practice in the health insurance domain, where numerous criteria need to be evaluated before a PA request is approved. Therefore, we use the criteria listed in \textit{Spine} guidelines of a major PA provider. Taking cost-constraints into account, we stratified sample the dataset to 500 data points from our larger set of 11,000 records. We used statistics like document length as well as made sure the ratio of categorical features is consistent between larger 11,000 records and its smaller 500 subset. Sub-sampling to such a small number could be made possible because the cardinality of categorical features is not high. The dataset has a mean document length of 7055 tokens, standard deviation of 8780, max of 804,759 while having $25^{th}$, $50^{th}$ and $75^{th}$ percentile of 2671, 4615 and 8386 tokens respectively.

\begin{table*}
\begin{center}
\begin{tabular}{p{0.2in}p{3.5in}p{0.5in}}
\toprule 

   & \textbf{Question} & \textbf{Answers} \\ \midrule
$Q_1$ & Are any of the conditions including Motor Weakness, Severe Radicular Pain, Cancer, Cauda Equina or Fracture present which could potentially be life or limb threatening? & Y, N\\
$Q_2$ & Is there a contraindication to MRI? & Y, N, NA\\
$Q_3$ & Has there been at least 6 weeks of provider-directed conservative treatment? & Y, N, NA\\
$Q_4$ & Was a clinical re-evaluation performed after a trial of conservative treatment? & Y, N, NA\\
$Q_5$ & Did symptoms improve with conservative treatment? & Y, N, NA\\
\bottomrule

\end{tabular}
\caption{Question-Answer (QA) pairs where answers Y, N, and NA stand for \textit{Yes}, \textit{No} and \textit{Not Applicable}}
\label{tab:qa}
\end{center}
\end{table*}

In total, there are 5 QA pairs (shown in Table \ref{tab:qa}) which are asked for all the case reports or data points. The answer choices for all these questions contain either of \textit{Yes}, \textit{No} or \textit{Not applicable}. The Yes and No choices are self-explanatory, however \textit{Not applicable} may need additional analysis. For example, for $Q_4$ and $Q_5$ if the \textit{conservative treatment} itself was not provided, we put them under the answer choice of \textit{Not applicable}.Q1

\section{Prompting Techniques}
We elaborate on different prompts that we use for GPT for our underlying QA use case. 

\begin{figure*}
\fbox{\begin{minipage}{38em}
You are a physician to review health record notes of a medical procedure request, then to choose the best answer for five muti-choice questions related to this request.    
 
\hfill\break
The list of multi-choice questions that need to be answered paired with their answer choices is listed below.

\hfill\break
Question: \textit{\{question\_text\_list\}}
\hfill\break
Answer Choices: {\textit{\{choices\_list\}}}

\hfill\break
Here are some health records notes from a doctor ordering diagnostic imaging for a patient.

\hfill\break
Health Records:
\#\#\#
\textit{\{clinical\_text\}}
\#\#\#

\end{minipage}}
\caption{Basic All Questions Prompt Template}
\label{fig:basic}
\end{figure*}
\paragraph{\textbf{Basic All Questions (Basic AQ)}}
\label{basic}

The prompt template used for this technique is shown in Figure \ref{fig:basic}. The Basic AQ approach asks GPT to answer all the 5 questions in one go. The \textit{question\_text\_list} placeholder specifies the list of all 5 questions and the placeholder \textit{choices\_list} specifies the corresponding answer choices. The \textit{clinical\_text} placeholder contains patient details which GPT needs to search through for the answers.

\begin{figure*}
\fbox{\begin{minipage}{38em}
You are a physician to review health record notes of a medical procedure request, then to choose the best answer for five muti-choice questions related to this request with the help of some medical term definitions. The relevant medical terms paired with definition are listed below.

\hfill\break
Medical Term Definitions:
\#\#\#
\textit{\{term\_definitions\}}
\#\#\#
 
\hfill\break
The list of multi-choice questions that need to be answered paired with their answer choices is listed below.

\hfill\break
Question: \textit{\{question\_text\_list\}}
\hfill\break
Answer Choices: {\textit{\{choices\_list\}}}

\hfill\break
Here are some health records notes from a doctor ordering diagnostic imaging for a patient.

\hfill\break
Health Records:
\#\#\#
\textit{\{clinical\_text\}}
\#\#\#

\end{minipage}}
\caption{Basic with Term Definitions All Questions Prompt Template}
\label{fig:basictd}
\end{figure*}
\paragraph{\textbf{Basic with Term Definitions All Questions (Basic TD + AQ)}}
\label{basic_term}
The prompt template used for this technique is shown in Figure \ref{fig:basictd}. The \textit{term\_definitions} placeholder specifies all the medical term definitions. This prompting technique builds upon Basic AQ by enriching the guidelines with definitions of keywords found in them. We worked with clinicians and used publicly available repositories\footnote{https://www.ncbi.nlm.nih.gov/pmc/}
to gather information that could either define or elaborate these keywords. For example, \textit{conservative treatment} in $Q_3$ and $Q_4$ encompasses things like physical therapy, resting, using muscle relaxants, etc. We hypothesis that by adding this information, we are helping GPT gain more context while answering the questions.

\paragraph{\textbf{Basic Per Question (Basic PQ)}}
\label{basic_perq}

In contrast to Basic AQ approach, this prompting technique asks one question at a time. Each question and it's response is now mutually exclusive from one another. The prompt template used is same as that of Basic AQ as shown in Figure \ref{fig:basic}.

\paragraph{\textbf{Basic with Term Definitions Per Question (Basic TD + PQ)}}
This method builds upon Basic TD + AQ by asking one question at a time with term definitions. The prompt template used is same as that of Basic TD + AQ as shown in Figure \ref{fig:basictd}.

\begin{figure*}
\fbox{\begin{minipage}{38em}
You are a physician to review health record notes of a medical procedure request, then to choose the best answer for the given multi-choice question related to this request.    

\hfill\break
The multi-choice question that needs to be answered paired with answer choices is listed below.

\hfill\break
Question: \textit{\{question\_text\}}
\hfill\break
Answer Choices: {\textit{\{choices\}}}

\hfill\break
Think step by step.

\hfill\break
Here are some health records notes from a doctor ordering diagnostic imaging for a patient.

\hfill\break
Health Records:
\#\#\#
\textit{\{clinical\_text\}}
\#\#\#

\end{minipage}}
\caption{Chain-of-Thought Reasoning Per Question Prompt Template}
\label{fig:cot}
\end{figure*}

\paragraph{\textbf{Chain-of-Thought Reasoning Per Question (CoT PQ)}}
The rationale behind using CoT technique in our work is that there may be multiple smaller questions that need to be answered first in order to conclude the answer of the final asked question. For example, one of the questions which we ask GPT is \textit{Has there been at least 6 weeks of provider-directed conservative treatment?} which can easily be split into 3 questions \textit{Has there been any conservative treatment?}, \textit{Was the treatment provider-directed?} and \textit{What was the duration of conservative treatment?}. The prompt template used for this technique is shown in Figure \ref{fig:cot}.
\begin{figure*}
\fbox{\begin{minipage}{38em}

You are a physician to review health record notes of a medical procedure request, then to choose the best answer for the given multi-choice question related to this request. When presented with the multi-choice question, generate similar question-answer pairs as examples from health records notes. Afterward, proceed to solve the initial multi-choice question.   

\hfill\break
The initial multi-choice question that needs to answered paired with answer choices is listed below. 

\hfill\break
Question: \textit{\{question\_text\}}
\hfill\break
Answer Choices: {\textit{\{choices\}}}

\hfill\break
Generate three examples of multi-choice question-answer pairs from health records notes that are similar to the initial multi-choice question. Your examples should be distinct from each other and from the initial multi-choice question.

\hfill\break
Now, choose the best answer for the initial multi-choice question.

\hfill\break
Here are some health records notes from a doctor ordering diagnostic imaging for a patient.

\hfill\break
Health Records:
\#\#\#
\textit{\{clinical\_text\}}
\#\#\#

\end{minipage}}
\caption{Analogical Reasoning Per Question Prompt Template}
\label{fig:ar}
\end{figure*}

\paragraph{\textbf{Analogical Reasoning Per Question (AR PQ)}}
Inspired from \cite{yasunaga2023large}, we designed our own analogical reasoning prompting by tweaking the prompt to fit to our problem statement. We do this because unlike general domain, GPT would not be able to recall patient level healthcare knowledge as it was never trained on one. Rather, we frame the prompt so that GPT doesn't need to rely a lot on global knowledge. 
To that end, instead of asking GPT to generate any kind of relevant QA pairs based on it's global knowledge, we ask GPT to generate QA pairs from the given patient records and then answer the question. We specifically ask GPT to generate three QA pairs. The prompt template used for this technique is shown in Figure \ref{fig:ar}.

\begin{figure*}
\fbox{\begin{minipage}{38em}

You are a physician to review health record notes of a medical procedure request, then to choose the best answer for the given multi-choice question related to this request. When presented with the multi-choice question, identify relevant sections or text extracts from health records notes which may help in answering the multi-choice question. Afterward, proceed to solve the given multi-choice question.   

\hfill\break
The multi-choice question that needs to be answered paired with answer choices is listed below.

\hfill\break
Question: \textit{\{question\_text\}}
\hfill\break
Answer Choices: {\textit{\{choices\}}}

\hfill\break
Identify three most relevant sections or text extracts from health records notes that may help in answering the multi-choice question. The identified sections or text extracts should be distinct from each other. The identified sections or text extracts must be between 50 to 200 words long.

\hfill\break
Now, choose the best answer for the given multi-choice question using the identified sections or text extracts.

\hfill\break
Here are some health records notes from a doctor ordering diagnostic imaging for a patient.

\hfill\break
Health Records:
\#\#\#
\textit{\{clinical\_text\}}
\#\#\#

\end{minipage}}
\caption{Implicit RAG Per Question Prompt Template}
\label{fig:irag}
\end{figure*}

\paragraph{\textbf{Implicit Retrieval Augmented Per Question (Implicit RAG PQ)}}
Most of the work on RAG talk about data retrieval based on an accepted relevancy score and then using LLM prompts to answer the given query. The data retrieval is done by storing the embeddings from some encoder of the entire corpus (one patient record in our case) in a vector database index and then retrieving the most matching data points (text extracts or sections from the patient records in our case) for a given query. The key rationale behind using RAG is that it helps in saving a lot of computational cost and improves the LLM's performance as now it has to look in a smaller knowledge space to answer the asked question. In our proposed novel prompting technique Implicit RAG, we completely ignore the overhead involved in the getting the embeddings of the entire corpus and storing them in a vector database. Instead, we ask the LLM itself to find the most relevant text extracts or sections in the given patient record which may help in answering the asked question and then later use these extracted sections to conclude to the answer of the original question. We specifically ask GPT to extract 3 relevant sections. The prompt template used for this technique is shown in Figure \ref{fig:irag}.

\section{Results \& Analysis}
We use the 32k context window version of GPT-4 to conduct all our experiments. We set the temperature, frequency penalty, presence penalty to be 0 and max tokens to be 1000. We set the temperature to be 0 so that we can reproduce the results of the GPT model. Setting frequency penalty and presence penalty to 0 ensures that there are no token-based restrictions on generations from the model. Since our dataset is highly imbalanced, we used weighted F1 as our evaluation metric. The results
for all the discussed approaches are shown in Table \ref{tab:wf1}. Overall, we find the our proposed approach improved the mean weighted F1-score by 3 points.

\begin{table*}
\centering
\begin{tabular}{lcccccc}
\toprule \textbf{Prompt} & \textbf{$Q_1$} & \textbf{$Q_2$} & \textbf{$Q_3$} & \textbf{$Q_4$} & \textbf{$Q_5$} & \textbf{Mean}  \\ \midrule
Basic All Questions\textsuperscript{*} & 0.74 & 0.77 & 0.49 & 0.61 & 0.60 & -\\
Basic with Term Definitions All Questions\textsuperscript{*} & 0.33 & 0.75 & 0.49 & 0.61 & 0.60 & - \\
Basic Per Question & 0.68 & 0.79 & 0.38 & 0.40 & \textbf{0.61} & 0.57 \\
Basic with Term Definitions Per Question & 0.48 & 0.78 & 0.40 & 0.39 & 0.60 & 0.53\\
Chain-of-Thought Per Question & \textbf{0.77} & \textbf{0.81} & 0.39 & 0.45 & 0.55 & 0.59 \\
Analogical Reasoning Per Question & 0.62 & 0.77 & 0.40 & 0.50 & 0.52 & 0.56 \\
Implicit RAG Per Question & 0.64 & 0.79 & \textbf{0.49} & \textbf{0.61} & 0.53 & \textbf{0.61} \\

\bottomrule
\end{tabular}
\caption{\label{tab:wf1} Weighted F1 for All Prompting Techniques. \textsuperscript{*} We don't calculate mean for Basic AQ and Basic TD + AQ or compare them with other methods as their results are biased. Please refer to \textbf{All Questions Vs Per Question} for more details.}

\end{table*}

\begin{table*}

\centering
\begin{tabular}{ccccccccccc}
\toprule
\textbf{Questions} &\multicolumn{2}{c}{$Q_1$ (20)} &\multicolumn{2}{c}{$Q_2$ (20)} &\multicolumn{2}{c}{$Q_3$ (20)} &\multicolumn{2}{c}{$Q_4$ (20)} &\multicolumn{2}{c}{$Q_5$ (20)}\\
\midrule
\textbf{Pattern} & \cmark (15) & \xmark (5) & \cmark (15) &
  \xmark (5) & \cmark (7) & \xmark (13) & \cmark (10) & \xmark (10) & \cmark (11) & \xmark (9) \\
\midrule

Right Section & 93\% & 100\%  & 100\% & 80\% & 86\% & 92\% & 90\% & 100\%  & 100\% & 100\% \\
Wrong Section & 7\% & 0\%  & 0\% & 20\% & 14\% & 8\%& 10\% & 0\%  & 0\% & 0\% \\

\bottomrule
\end{tabular}
\caption{Qualitative Analysis of \textit{Implicit RAG} on $Q_1$, $Q_2$, $Q_3$, $Q_4$ and $Q_5$}
\label{compare}
\end{table*}

\paragraph{\textbf{All Questions Vs Per Question}}
\label{aq_pq}
The numbers in Table \ref{tab:wf1} seem to convey that the Basic AQ and the Basic TD + AQ outperform all the other techniques but there is a catch here. As we mentioned earlier, in our study we use a dataset of 500 data points. Therefore, when calculating evaluation metrics for all the prompts, our support remains 500 except for Basic AQ and Basic TD + AQ. For these two prompting techniques, the support on average is just 421. GPT is unable to generate structured output for 79 documents and hence the results are biased. This hints that increasing the number of questions in a single prompt leads to either inability to answer at all or to generate structured output as instructed in the prompt.

\paragraph{\textbf{Question Wise Analysis}}

\textit{\textit{Implicit RAG}} is able to outperform all other discussed techniques. It obtains the best weighted scores for $Q_3$ and $Q_4$ and comparable scores for $Q_1$ and $Q_5$. For $Q_1$, CoT significantly outperforms other techniques. It could be because of the nature of the question which requires breaking down of the question into multiple smaller questions with respect to checking all the different kinds of life or limb-threatening condition possible like Cancer, Fracture, etc. For $Q_2$ and $Q_5$, all the techniques perform similarly. \textit{Implicit RAG} outperforms all other methods for $Q_3$ and $Q_4$. Both $Q_3$ and $Q_4$ involve a temporal factor that can be spread in many different segments throughout the lengthy documents. Hence, by using \textit{Implicit RAG}, GPT is able to solve this problem incrementally by first identifying the relevant textual extracts and then answering the question. 

\paragraph{\textbf{Do Term Definitions Help in Anyway?}}
Looking at the results of Basic AQ, Basic TD + AQ, Basic PQ and Basic TD + PQ in Table \ref{tab:wf1}, we can say that the additional context provided by medical term definitions do not really make a difference for questions $Q_1$, $Q_3$, $Q_4$ and $Q_5$. For $Q_1$, the term definitions on the contrary bring down the performance. It could be possible that the term definitions used in $Q_1$ are very narrow in their knowledge scope and conflict with the bigger knowledge base that GPT inherently carries. For example, $Q_1$ talks about a lot of intricate medical terms like Cancer which is a big research field in itself and hence a small textual definition elaboration in the prompt ends up confusing GPT even more. Other questions apart from $Q_1$ do not really have a lot of intricate medical terms like $Q_1$.

\paragraph{\textbf{Implicit RAG}}
We do a qualitative analysis on 20 randomly picked patient health records for all the 5 questions for \textit{Implicit RAG}. We check how many times the extracted sections are relevant to the question or not. Even if 1 out of 3 extracted sections are relevant, we consider that to be a valid retrieval irrespective of whether the final answer was correct or incorrect. The results are shown in Table \ref{compare}. We calculate individual question scores across all 20 health records. As we can see, \textit{Implicit RAG} is able to extract relevant sections in most of the cases. 

\section{Ethical Statement}

Before using this dataset for our experiments, we took special care to de-identify all patient cases. De-identification of data, especially data containing Protected Health Information (PHI), is a critical step to protect privacy and comply with data regulations like Health Insurance Portability and Accountability Act (HIPAA). De-identification involves removing or altering any information that could potentially identify individuals while retaining the utility of the data for research, analysis, or other purposes. Different techniques like obfuscation with masking of protected entities and replacement with pseudo entities were done for de-identification. Again, due to data regulations, this dataset is not publicly available.


\section{Conclusion \& Future Work}
In this work, we study different prompting techniques and how they can help in addressing the decisions associated with a PA approval via a guideline based QA task. We also come with a novel prompting method \textit{Implicit RAG} which first fetches textual segments from the patient record before reasoning over them to answer the asked question. The main idea behind using \textit{Implicit RAG} is that it helps the LLM in shrinking its knowledge base that it needs to refer to while answering a question. 

One of the directions of the future work could be to develop an ensemble prompting technique which combines the strengths of different techniques like CoT, AR, \textit{Implicit RAG}. In this ensemble approach, we can have a weighing mechanism while selecting the output of different prompting methods based on the nature of the question asked. For example, if the question deals with temporal aspect, probably \textit{Implicit RAG's} output could be given the maximum weight. Similarly, if the question can be divided into smaller questions, CoT's output gets the maximum weight.
Another direction of future work could be to apply techniques which can identify relevant portions of the unconventionally big patient records where the probability of finding the answer to the asked guideline question is maximum. This will not only reduce the processing time by the model when trying to find the answers and but will also help in consuming input documents which exceed pre-defined token limit or context window size of the model. Although the recent LLMs have been continuously working towards increasing their context window size (latest GPT-4 Turbo has 128k context window size), there will always be patient records exceeding LLM's pre-defined context window size and hence the above-mentioned approach could help in such scenarios.

\bibliographystyle{author/spphys}
\bibliography{author/ref}
\end{document}